\documentclass[letterpaper]{article} 
\usepackage{aaai25}  
\usepackage{times}  
\usepackage{helvet}  
\usepackage{courier}  
\usepackage[hyphens]{url}  
\usepackage{graphicx} 
\usepackage{natbib}  
\usepackage{caption} 
\usepackage{algorithm}
\usepackage{algorithmic}
\usepackage{amssymb}
\usepackage{multirow}
\usepackage{xcolor}
\usepackage{colortbl}
\usepackage{enumitem}
\usepackage{amsmath}
\usepackage{amsfonts}
\usepackage{booktabs}
\definecolor{lightblue}{rgb}{0.93,0.95,1.0}

\usepackage{newfloat}
\usepackage{listings}
\DeclareCaptionStyle{ruled}{labelfont=normalfont,labelsep=colon,strut=off} 
\floatstyle{ruled}
\newfloat{listing}{tb}{lst}{}
\floatname{listing}{Listing}
\title{MindTuner: \\Cross-Subject Visual Decoding with Visual Fingerprint and Semantic Correction}
\author{
    Zixuan Gong\textsuperscript{\rm 1},
    Qi Zhang\textsuperscript{\rm 1}, 
    Guangyin Bao\textsuperscript{\rm 1}, \\
    Lei Zhu\textsuperscript{\rm 1}, 
    Rongtao Xu\textsuperscript{\rm 2}, 
    Ke Liu\textsuperscript{\rm 3}, 
    Liang Hu\textsuperscript{\rm 1}, 
    Duoqian Miao\textsuperscript{\rm 1,}\thanks{Corresponding author}
}
\affiliations{
    \textsuperscript{\rm 1}Tongji University, \textsuperscript{\rm 2}Chinese Academy of Sciences, \textsuperscript{\rm 3}Beijing Anding Hospital \\
    \texttt{\{gongzx,zhangqi\_cs,baogy,dqmiao\}@tongji.edu.cn}

}

\usepackage{bibentry}
\let\oldtwocolumn\twocolumn
\renewcommand\twocolumn[1][]{%
    \oldtwocolumn[{#1}{
    \begin{center}
    \vspace{-0.95cm}
    \includegraphics[width=\textwidth]{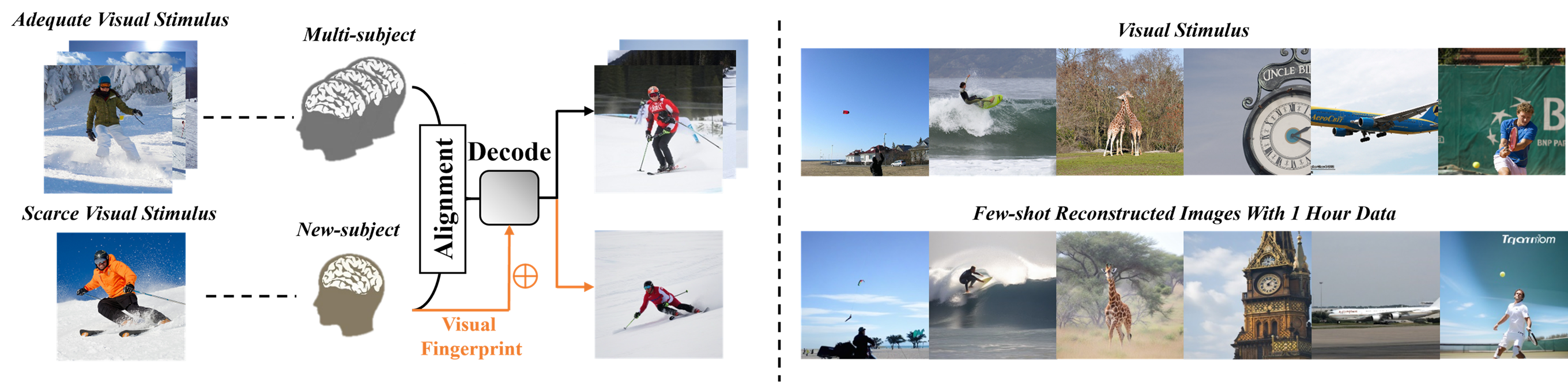}
    \captionof{figure}{Cross-Subject Visual Decoding and Image Reconstruction. Subjects with adequate fMRI data are aligned to decode visual stimuli via learning a shared network. A new subject, even with scarce visual stimulus, is aligned to the common space of the shared network, which perceives the subject's unique visual fingerprint to ensure precise visual decoding.
    }
    \label{1 hour}
    \end{center}
    }]
}

\begin{document}

\maketitle

\begin{abstract}
Decoding natural visual scenes from brain activity has flourished, with extensive research in single-subject tasks and, however, less in cross-subject tasks. Reconstructing high-quality images in cross-subject tasks is a challenging problem due to profound individual differences between subjects and the scarcity of data annotation. In this work, we proposed MindTuner for cross-subject visual decoding, which achieves high-quality and rich semantic reconstructions using only 1 hour of fMRI training data benefiting from the phenomena of visual fingerprint in the human visual system and a novel fMRI-to-text alignment paradigm. Firstly, we pre-train a multi-subject model among 7 subjects and fine-tune it with scarce data on new subjects, where LoRAs with Skip-LoRAs are utilized to learn the visual fingerprint. Then, we take the image modality as the intermediate pivot modality to achieve fMRI-to-text alignment, which achieves impressive fMRI-to-text retrieval performance and corrects fMRI-to-image reconstruction with fine-tuned semantics. The results of both qualitative and quantitative analyses demonstrate that MindTuner surpasses state-of-the-art cross-subject visual decoding models on the Natural Scenes Dataset (NSD), whether using training data of 1 hour or 40 hours.
\end{abstract}

\section{Introduction}
Do our brains form unified perceptions when we observe similar objects? Do our unique understandings influence these perceptions differently? The human brain exhibits substantial anatomical similarities in terms of functional organization, including shared attributes like memory, functional connectivity, and visual cortex functions~\cite{nn_shared,fingelkurts2005functional,visual2019nature}. However, individual neural biases always exist due to inherent differences~\cite{wang2020visualfingerprint}. Understanding both the similarities and gaps in perception has profound implications for the fields of Artificial Intelligence (AI)~\cite{BMI-Net,fMRI_auditory} and Brain-Computer Interface (BCI) research~\cite{brain_state_decoding}. Visual decoding is a straightforward way to understand the brain, where functional magnetic resonance imaging (fMRI) is a widely embraced non-invasive tool used to decode natural visual stimuli, revealing intricate perceptual and semantic details in the cerebral cortex~\cite{fMRIbook}. Consequently, fMRI has garnered considerable attention in image retrieval and reconstruction tasks.

Open-source large-scale fMRI datasets, such as the Natural Scenes Dataset (NSD)~\cite{nsd}, advance deep learning models to shine in fMRI decoding. Pre-trained cross-modality models like CLIP~\cite{clip} and Stable Diffusion~\cite{sd} offer effective representation space and models for high-quality visual reconstruction. A large body of literature demonstrates the feasibility of training single-subject decoding models to reconstruct high-fidelity images~\cite{mindreader2022,mindseye2023,chen2023seeing, takagi2023high,lu2023minddiffuser}. However, single-subject decoding has drawbacks, including the need to train a unique model for each subject, making it challenging to generalize to new subjects and requiring a substantial amount of fMRI data for training. As is widely recognized, acquiring a large amount of fMRI data for each subject is time-consuming, labor-intensive, and impractical in practical scenarios. Unfortunately, most current research focuses on single-subject visual decoding rather than exploring the challenging commonalities of the human brain. Consequently, there is an urgent need for cross-subject decoding models that can be effectively transferred to new subjects and perform well in the few-shot setting, as depicted in Figure \ref{1 hour}.

The key to cross-subject few-shot decoding lies in effectively utilizing extensive prior knowledge from other subjects or additional modalities. On the one hand, one successful strategy for leveraging knowledge from other subjects involves aligning them to a shared space. Ridge regression is commonly employed for this purpose, aligning voxel inputs from different subjects~\cite{scotti2024mindeye2,ferrante2023througheyes}. This approach is preferred due to the low signal-to-noise ratio in fMRI data, where complex non-linear models tend to overfit noise. Nonetheless, the process of visual information perception and generating brain activity in each individual incorporates unique components, referred to as the visual fingerprint~\cite{wang2020visualfingerprint} (refer to Preliminary for more detailed analysis). Current linear alignment methods only enable new subjects to conform to the shared components across subjects, neglecting the perception difference derived from their distinctive visual fingerprint and resulting in limited performance.

On the other hand, an additional strategy involves leveraging multimodal data. Previous alignment methods focused solely on the visual modality, 
as a means of adapting to the inputs of Stable Diffusion. However, this alignment approach is susceptible to slight disturbances, leading to semantic errors in the generated images. For visual decoding tasks, the textual modality is highly relevant and is verified effective in enhancing visual decoding semantically~\cite{scotti2024mindeye2}. However, previous approaches incorporating text have placed excessive emphasis on directly aligning fMRI with detailed textual descriptions to facilitate the reconstruction process~\cite{takagi2023high}, which intuitively lacks rationality and yields poor performance. Considering that subjects in visual stimulation experiments lack direct interaction with the textual modality and that individual understanding of visual stimuli varies, the relationship between fMRI and text should be considered implicit.

In this paper, we propose \textbf{MindTuner}: a cross-subject visual decoding framework. Inspired by visual fingerprint in Brain Science~\cite{wang2020visualfingerprint}, and with the help of Low-Rank Adaptation (LoRA) for large model lightweight fine-tuning. In correspondence to the above two strategies, we first propose the combination of non-linear \textbf{Skip-LoRAs} and \textbf{LoRAs} to learn the visual fingerprint of new subjects, which are injected into the fMRI encoding network to correct visual perception difference. In addition, we design a \textbf{Pivot} module that uses images as the central modality to bridge fMRI and text. The Pivot helps to correct the reconstructed image with fine-tuned semantics. Our contributions are summarized as follows:

\begin{itemize}[leftmargin=*]
\item MindTuner makes the first attempt to introduce \textbf{LoRA} as a subject-level fine-tuning module in cross-subject decoding and further elaborately design non-linear \textbf{Skip-LoRA}. Their combination shows excellent capability to learn subjects' visual fingerprint.
\item We introduce a novel fMRI-to-text retrieval paradigm with a \textbf{Pivot} using the image modality. The Pivot conducts semantic correction with label prompts to enhance fMRI-to-image reconstruction.
\item We evaluate our method on the NSD dataset, and it establishes SOTA decoding performance whether using training data of 40 hours (full) or 1 hour(2.5\% of full).
\end{itemize}

\section{Related Work}
\subsection{Cross-Subject Functional Alignment}
In visual decoding, single-subject models have exposed the issue of excessive reliance on the data volume of individual subjects, leading researchers to shift toward cross-subject studies.
Functional alignment of different brains is considered to be more effective than anatomical alignment. Previous functional alignment methods were mainly divided into two perspectives: fMRI data itself and downstream tasks. Among a series of methods starting from the fMRI perspective, Bazeille et al.~\cite{bazeille2019earlyot} and Thual et al.~\cite{thual2023metaaicross} minimize an optimal transport cost between voxels of different brains. The methods based on fMRI self-supervision~\cite{chen2023seeing,qian2023fmri-pte} emphasize obtaining common latent representations between subjects through autoencoders. Wills Aligner ~\cite{bao2024wills} aligns data from different subjects using the fsaverage template. On another cross-subject alignment research path, most of these methods focus on the shared knowledge among subjects and overlook the complex nonlinear relationships between subjects due to concerns about overfitting. Linear fitting ~\cite{ferrante2023througheyes} was performed on the responses of subjects to common images, achieving high-quality cross-subject visual reconstruction, but requiring subjects to see the same images. Mindeye2 ~\cite{scotti2024mindeye2}, MindBridge ~\cite{wang2024mindbridge} and UMBRAE ~\cite{xia2024umbrae} improved this and achieved impressive results, but still losing subject-specific features and ignoring the more complex relationships between subjects, which is not flexible.

\subsection{Text Modality in Visual Decoding}
Decoding visual stimuli from fMRI has been a long-standing endeavor, primarily focusing on the image modality ~\cite{takagi2023high,lu2023minddiffuser,litemind,brain-diffuser, mindseye2023}. However, an increasing number of studies have highlighted the role of text.  UniBrain~\cite{mai2023unibrain} directly aligns fMRI to the text representation via ridge regression and then completes the brain caption task with the text generation model. MindEye2~\cite{scotti2024mindeye2} obtains a better caption by placing the generated image representation into the image captioning model, which is used to smooth the generated image. NeuroClips utilizes the pretrained BLIP-2 ~\cite{li2023blip} model to achieve text-assisted video reconstruction ~\cite{gong2024neuroclips}. However, these methods either do direct fMRI-to-text alignment or direct image-text alignment, ignoring the indirect relationship between fMRI and text from the perspective of intuitive understanding. 

\section{Preliminary on visual fingerprint}
\label{sec:pre}
\begin{figure}
  \centering
   \includegraphics[width=1.0\linewidth]{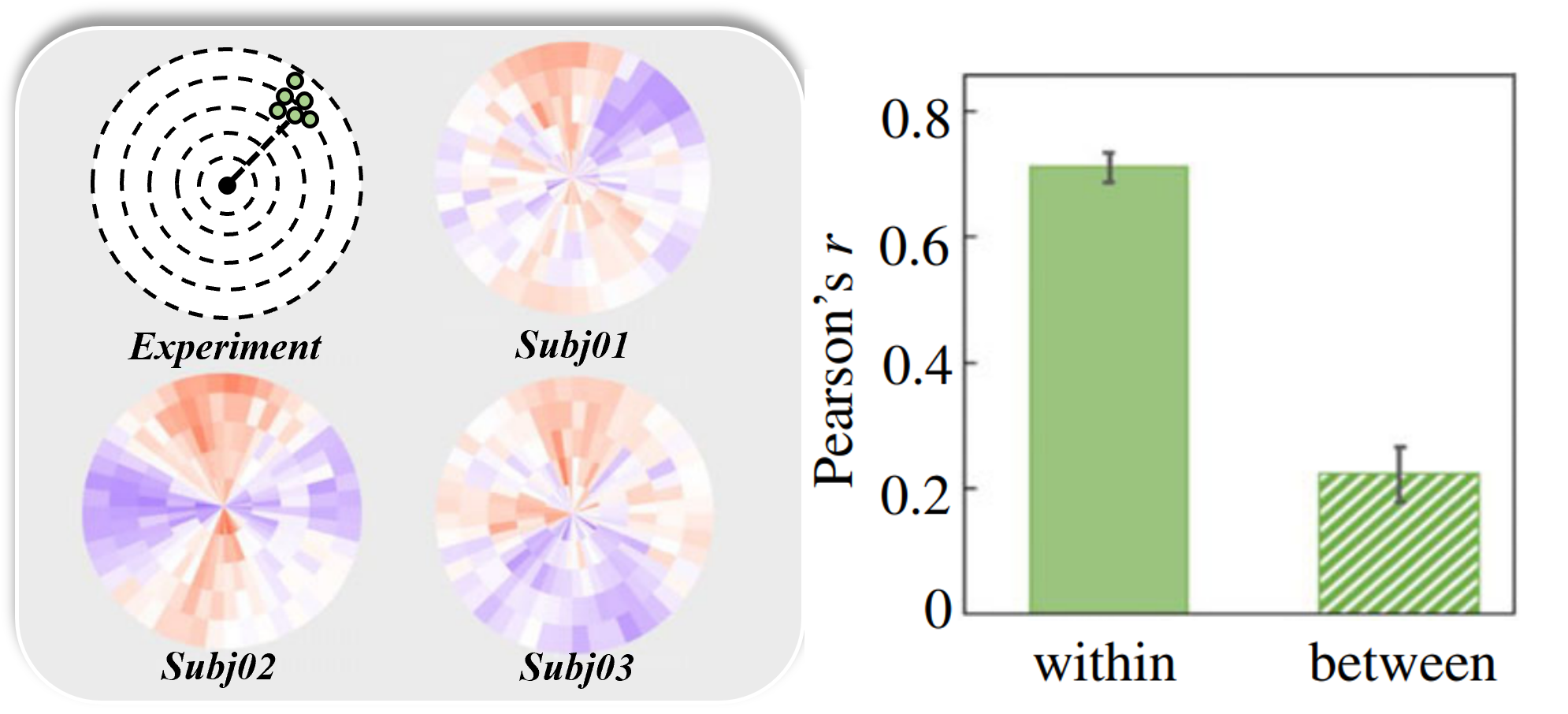}
   \vspace{-0.6cm}
   \caption{Visual fingerprint experiments across subjects. 'Within' denotes Pearson correlation coefficient of Distortion Indices in within-subject experiments, while 'between' denotes between-subject.}
   \label{visual fingerprint}
   \vspace{-0.5cm}
\end{figure}

Visual stimuli are processed by the Human Visual System(HVS). 
Therefore, as stated in previous research~\cite{xia2024dream}, brain responses such as fMRI are closely linked to our visual system. Although presented in a systematic manner, there are still significant differences in the visual systems of different individuals. Research of visual fingerprint emphasizes that idiosyncratic biases exist in the underlying representation of visual space propagate across varying levels of visual processing~\cite{wang2020visualfingerprint}. Using a position-matching task will find stable subject-specific compressions and expansions within local regions throughout the visual field. As shown on the left of Figure \ref{visual fingerprint}, in experiments, subjects were fixated at the center, and a target was displayed briefly at one of five possible eccentricities (depicted by dashed lines, which were not visible in the experiment). After the target disappeared, subjects moved the cursor to match the target’s location. Linear models were used to fit the Distortion Indices(DI), which measured the degree of spatial distortion between subjects:
\begin{equation}
\begin{aligned}
    DI_{self}&\sim\beta_0 + \beta_1 \times self,\\
    DI_{others}&\sim\beta'_0 + \beta'_1 \times others.\end{aligned}
\end{equation}
As shown on the right of Figure \ref{visual fingerprint}, the experiment results show that subjects have their own visual fingerprint, with both linear and non-linear components. Within-subject similarity (r=0.71) is significantly higher than between-subject similarity (r=0.22), suggesting that each individual subject has their own unique spatial distortions that are consistent within themselves and distinguished from others.

\section{Method}
\begin{figure*}
  \centering
   \includegraphics[width=1.0\linewidth]{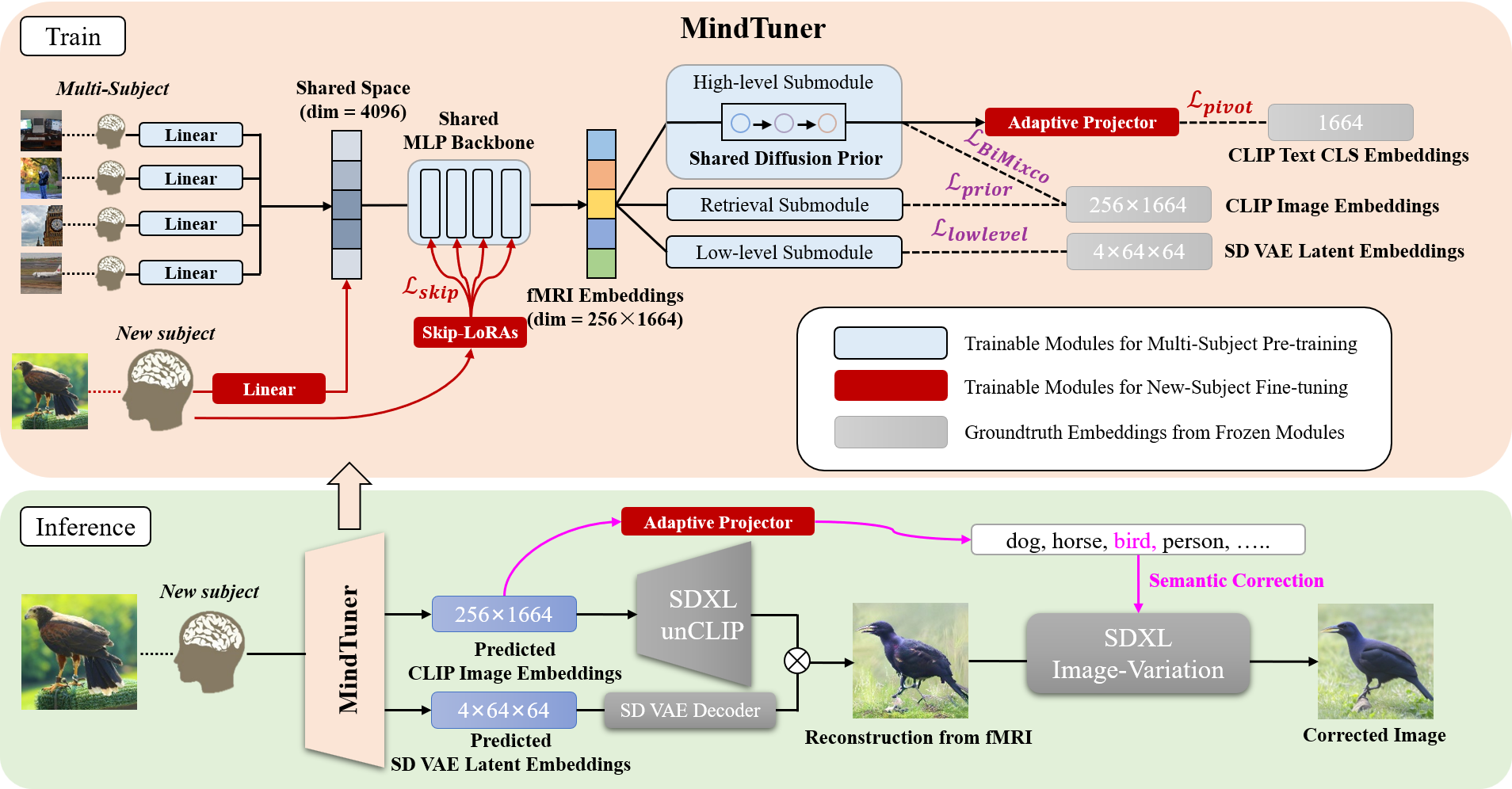}
   \vspace{-0.6cm}
   \caption{Schematic diagram of MindTuner. The training process was split into two phases: Multi-subject Pre-training and Cross-subject Fine-tuning, in which the corresponding modules were trained. The predicted embeddings are first obtained through MindTuner, and then the preliminary reconstructed image was obtained by SDXL unCLIP. The final reconstructed image is obtained by text retrieval and semantic correction by SDXL Image-Variation.}
   \label{model}
\end{figure*}
The task of cross-subject visual decoding consists of two parts: multi-subject pre-training and new-subject fine-tuning. Given a neural dataset with brain activities of subject $s$, it involves the reconstruction of visual images $I \in \mathbb{R}^{3 \times H \times W}$ through a core pipeline whose inputs are flattened and aligned fMRI voxels $V \in \mathbb{R}^{1 \times d_s}$ after ROI extraction, where $d_s$ denotes voxel numbers. To simplify notation, each $(I, V)$ denotes data from the original subjects for this section. Our goal is to optimize the $\mathcal{F(\cdot)}$, so that $\mathcal{F}(V)=\hat{I}$, where $\hat{I}$ best approximates $I$. New-subject fine-tuning adheres to the same path, albeit with scarce data. For the multi-subject pre-training, we follow the pipeline of MindEye2
, while for the new-subject fine-tuning, we plug into visual fingerprint and Pivot as shown in Figure \ref{model}.

\subsection{Multi-Subject Pre-training}
\label{sec4.1}

\subsubsection{Multi-Subject Functional Alignment}
It should be noted that the brain structures of different subjects vary, as do the number of voxels obtained. Consequently, a mapping model is required to align the voxel inputs from different subjects to the same dimension as the inputs to the model in the phase of multi-subject pre-training. 
Here, we employ linear function alignment to learn shared-subject fMRI latent space $M\in\mathbb{R}^{1\times d_0}$($d_0$ denotes shared input dimension). This is achieved through subject-specific ridge regression, as detailed below:
\begin{equation}
    M = \textbf{Ridge}^{(s)}(V)
\end{equation}

\subsubsection{MLP Backbone}
In order to achieve high-fidelity reconstruction, it is necessary to utilize a substantial quantity of CLIP image embedding. The OpenCLIP ViT-bigG/14 space is employed for alignment, with an image embedding dimension of $256\times1664$. Mapped inputs $M$ are fed into an MLP backbone comprising four residual blocks and a tokenization layer which transforms the input from a dimension of $d_0$ to $256\times1664$. The MLP Backbone $\epsilon(\cdot)$ serves to convert the fMRI to the intermediate backbone embedding space: $Z = \epsilon(M)$. It should be noted that all subjects shared the same MLP backbone following multi-subject functional alignment. Subsequently, the backbone embeddings are conveyed into three submodules for retrieval, high-level reconstruction, and low-level reconstruction.

\subsubsection{Retrieval Submodules}
A straightforward approach to performing the retrieval task is to conduct a shallow mapping of the backbone embeddings and supervise it with an fMRI-to-image CLIP contrastive loss. In the case of limited fMRI data, the application of appropriate data augmentation techniques can facilitate the convergence of the model. A recently proposed voxel mixture paradigm, based on MixCo ~\cite{kim2020mixco}, has demonstrated effectiveness. Two raw fMRI voxels $V_i$ and $V_j$ are mixed into $V_{mix_{i,j}}$ using a factor $\lambda$ sampled from the Beta distribution:
\begin{equation}
\begin{aligned}
    V_{mix_{i,j}} &= \lambda_i V_i + (1-\lambda_i) V_j, \\
    M_{mix_{i,j}} &= \textbf{Ridge}^{(s)} ( V_{mix_{i,j}} ), \\
    Z_{mix_{i,j}} &= \epsilon(M_{mix_{i,j}}) \\
\end{aligned}
\end{equation}
where $j$ denotes an arbitrary mixing index in the batch. The forward mixed contrastive loss MixCo is formulated as:
\begin{equation}
\begin{aligned}
    \mathcal{L}_{MixCo}=-\frac{1}{|B|}\sum_{i=1}^{|B|} [\lambda_i\log \frac{\exp(Z_{mix_{i,j}}\cdot\ g_i/{\tau})}{\sum_{m=1}^{|B|} \exp(Z_{mix_{i,j}}\cdot\ g_m/{\tau})} \\
    + (1-\lambda_i)\log \frac{exp(Z_{mix_{i,j}}\cdot\ g_j/{\tau})}{\sum_{m=1}^{|B|} \exp(Z_{mix_{i,j}}\cdot g_m/{\tau})}]
\end{aligned}
\end{equation}
where $g$ denotes ground-truth CLIP image embeddings, $\tau$ denotes a temperature hyperparameter, and $B$ is the batch size. Here we use the bidirectional loss $\mathcal{L}_{BiMixCo}$.
\subsubsection{Low-level and High-level Submodules}
The low-level pipeline is a widely utilized technique for the enhancement of low-level visual metrics in reconstruction images. This involves the mapping of voxels to the latent space of Stable Diffusion's Variational AutoEncoder (SDVAE), which serves as a surrogate for the reconstruction. The pipeline comprises an MLP and a CNN upsampler with L1 loss in Stable Diffusion's latent embeddings $z$.
\begin{equation}
    \mathcal{L}_{lowlevel} = \frac{1}{|B|}\sum_{i=1}^{|B|}|z_i-\hat{z_i}|
\end{equation}
Conversely, a high-level pipeline places greater emphasis on semantic alignment. Inspired by DALLE·2~\cite{ramesh2022dalle2}, a diffusion prior is recognized as an effective means of transforming backbone embeddings into CLIP ViT image embeddings, in which mean square error loss is used (further insights on the deployment of diffusion priors can be found in Appendix B.4):
\begin{equation}
    \mathcal{L}_{prior} = \frac{1}{|B|}\sum_{i=1}^{|B|}||g_i-\hat{g_i}||^2_2
\end{equation}
Thus, the end-to-end loss for multi-subject pre-training is:
\begin{equation}
    \mathcal{L}_{multi} = \mathcal{L}_{prior}+\alpha_1\mathcal{L}_{lowlevel} + \alpha_2\mathcal{L}_{BiMixCo}
\end{equation}
Where $\alpha$ denotes the weight to balance multiple losses.
\subsection{New-Subject Fine-tuning}
\label{sec4.2}
\subsubsection{Low-Rank Adaptation}
Previous work has demonstrated the effectiveness of low-rank adaptation in fine-tuning large language models with significantly fewer trainable parameters. It is well suited to our multi-subject decoding task for two reasons. First, most of the current mainstream models for fMRI decoding are MLP-based models that contain a large number of linear layers, whereas LoRA has been shown to achieve good fine-tuning results in the linear layers. Second, in cross-subject scenarios, fMRI data from new subjects are usually scarce, and full-tuning the whole model is usually difficult to grasp, leading to a certain degree of overfitting.
For each pre-trained weight matrix $\in\mathbb{R}^{d_{in} \times d_{out}}$ in the multi-subject model, where $d_{in}$ denotes input dimension and $d_{out}$ denotes output dimension, gradient update is constrained with a low-rank decomposition for new-subject adapter matrix $\Delta W$:
\begin{equation}
    W+\Delta W=W+BA
\end{equation}
where $W$ is keeped frozen, $B\in\mathbb{R}^{d_{out}\times r}$, $A\in\mathbb{R}^{r\times d_{in}}$, $r$ is the rank and $r\ll min(d_{in},d_{out})$. At the beginning of the training phase, the parameters of the matrix $B$ are randomly initialized, and $A$ is initialized to zero, which ensures that the initial output of the LoRA block is all zeros. 
\subsubsection{non-linear Skip-LoRAs}
The LoRA model is notably lightweight, effectively circumvents complexity, and could avoid the aforementioned fMRI overfitting problem. Although LoRA can be an effective means of fine-tuning multi-subject models, simple linear LoRA models are insufficient for the capture of visual fingerprints. Indeed, there exists a non-linear relationship between subjects (discussed in the Preliminary Section). Therefore, we inject a non-linear design into it. Here, we design our non-linear LoRAs by adding the activation function and the nonlinear constraints. Inspired by the skip-connection of Unet in Computer Vision~\cite{controlnet}, we built a brand new Skip-LoRAs to bootstrap the initial non-linearity of fMRI between subjects directly affecting the entire MLP backbone(see Appendix B.2 for Skip-LoRAs). Skip-LoRAs can be defined as $\textbf{Skip-LoRA} = Activation(BA)$. Assuming that $\epsilon_h(\cdot)$ is the $h$-th layer of MLP backbone $\epsilon(\cdot)$, the output of the new-subject backbone $M_h$ could be as follows:
\begin{equation}
\begin{aligned}
    &M_0 = \textbf{Ridge}^{(s)} (V),  \\
    &M_h = M_{h-skip} + M_{h-linear},  \\
    &M_{h-linear} = \epsilon_h(M_{h-1})+\textbf{LoRA}_h(M_{h-1}), \\
    &M_{h-skip} = \textbf{Skip-LoRA}_h(V) \\
\end{aligned}
\end{equation}
Here, we use the Pearson correlation coefficient to define the non-linear correlation loss:
\begin{equation}
    \mathcal{L}_{skip} = \frac{1}{|B|}\frac{1}{|H|}\sum_{i=1}^{|B|}\sum_{h=1}^{|H|}|Pearson(M_{h-linear},M_{h-skip})|
\end{equation}
\subsubsection{Pivot with Adaptive Projector}
In addition to pixel-level alignment, semantic-level matching is equally important for reconstructing the semantic integrity of an image. In visual decoding tasks, fMRI is directly associated with the image stimulus, and semantic information is implicit in fMRI. More details on the Adaptive Projector can be found in Appendix B.3. For new subjects, we add the additional image-text loss to the model to constrain the semantic information:
\begin{equation}
    \mathcal{L}_{pivot} = -\frac{1}{|B|}\sum_{i=1}^{|B|} \log \frac{\exp(p_i\cdot\ t_i/{\tau})}{\sum_{m=1}^{|B|} \exp(p_i\cdot\ t_m/{\tau})}
\end{equation}
where $p$ denotes the projector's output from image tokens $g$, and $t$ denotes the CLS embeddings (CLS denotes the classification token) of paired text. During new-subject fine-tuning, the adaptive projector is trainable. 
Thus, the end-to-end loss for new-subject fine-tuning is:
\begin{equation}
    \mathcal{L}_{new} = \mathcal{L}_{multi} + \alpha_3\mathcal{L}_{skip}+\alpha_4\mathcal{L}_{pivot}
\end{equation}
\subsubsection{Semantic Correction}
\begin{figure}[!t]
  \centering
   \includegraphics[width=1.0\linewidth]{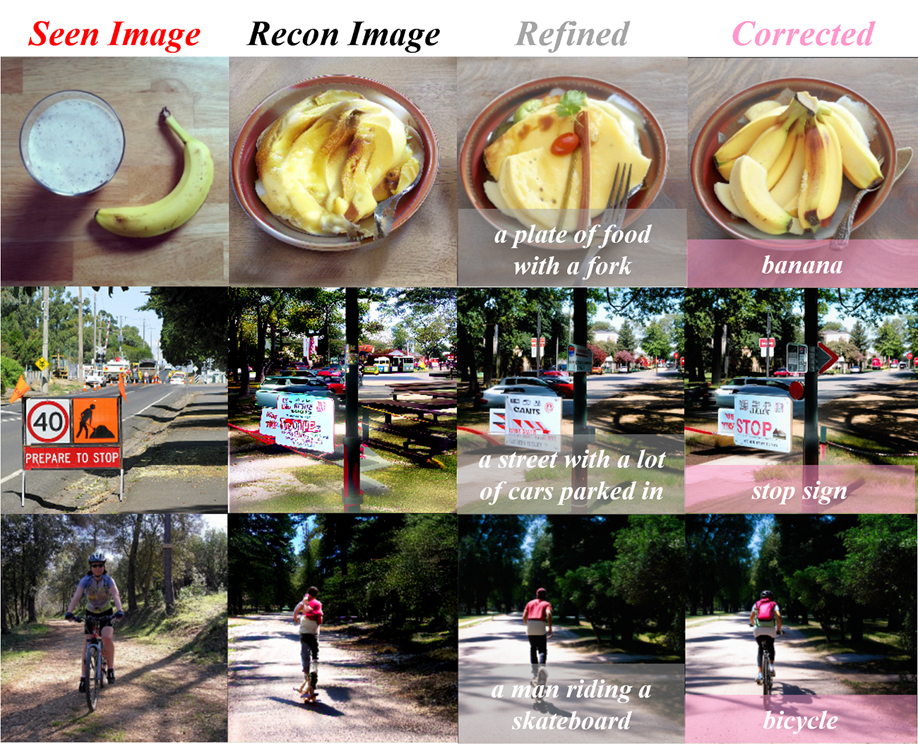}
   \vspace{-0.6cm}
   \caption{SDXL unCLIP reconstructions and SDXL Image-Variation by MindEye2's refinement or our correction.}
   \vspace{-2mm}
   \label{class correction}
\end{figure}

MindEye2 has found that the reconstructed images from SDXL unCLIP after token alignment have fuzzy semantics, so the aligned embeddings are fed into the image captioning model GIT~\cite{wang2022git} to get the semantics for refinement. We have found that single category words are sufficient to accomplish the refinement task. In addition, the captions generated by MindEye2 are completely dependent on aligned embeddings, which can only make the image semantics clearer and cannot correct the semantics of the reconstructed images, as shown in Figure \ref{class correction}. Taking advantage of our adaptive projector, we can define category description as an fMRI-to-text retrieval task using a simple text prompt $[class]$. In this way, we achieve more accurate category reconstruction.
\begin{table*}[!t]
  \centering
  \begin{tabular}{@{}ccccccccccc@{}}
    \toprule
    \multirow{2}{*}{Method} & \multicolumn{4}{c}{Low-Level}& \multicolumn{4}{c}{High-Level}& \multicolumn{2}{c}{Retrieval}\\
    \cmidrule(lr){2-5} \cmidrule(lr){6-9} \cmidrule(lr){10-11}
     &PixCorr$\uparrow$&SSIM$\uparrow$&Alex(2)$\uparrow$&Alex(5)$\uparrow$&Incep$\uparrow$&CLIP$\uparrow$&Eff$\downarrow$&SwAV$\downarrow$&Image$\uparrow$&Brain$\uparrow$\\

    \midrule
     Takagi...&0.246&0.410&78.9\%&85.6\%&83.8\%&82.1\%&0.811&0.504&-&-\\
     Ozcelik...&0.273&0.365&94.4\%&96.6\%&91.3\%&90.9\%&0.728&0.422&18.8\%&26.3\%\\
     MindEye1&\underline{0.319}&0.360&92.8\%&96.9\%&94.6\%&93.3\%&0.648&0.377&90.0\%&\underline{84.1\%}\\
     MindEye2&\textbf{0.322}&\textbf{0.431}&\textbf{96.1\%}&\underline{98.6\%}&\underline{95.4\%}&\underline{93.0\%}&\underline{0.619}&\underline{0.344}&\underline{98.8\%}&\textbf{98.3\%}\\
     \textbf{MindTuner}(Ours)&\textbf{0.322}&\underline{0.421}&\underline{95.8\%}&\textbf{98.8\%}&\textbf{95.6\%}&\textbf{93.8\%}&\textbf{0.612}&\textbf{0.340}&\textbf{98.9\%}&\textbf{98.3\%}\\
     \cmidrule(lr){1-11}
     MindEye2(1 hour)&0.195&0.419&84.2\%&90.6\%&81.2\%&79.2\%&0.810&0.468&79.0\%&57.4\%\\
     \textbf{MindTuner}(1 hour)&\textbf{0.224}&\textbf{0.420}&\textbf{87.8\%}&\textbf{93.6\%}&\textbf{84.8}\%&\textbf{83.5}\%&\textbf{0.780}&\textbf{0.440}&\textbf{83.1\%}&\textbf{76.0\%}\\
     \bottomrule
  \end{tabular}
  \vspace{-0.1cm}
  \caption{Quantitative comparison of MindTuner's performance against other methods. All results are averaged across subjects 1, 2, 5, and 7 from the Natural Scenes Dataset. The results of other methods are taken from MindEye2\cite{scotti2024mindeye2}, using either 40-hour data or 1-hour data. Some methods were not included in the comparison because they used past 37-hour data or were not open-source. Missing values occur when metrics are not applicable. Bold font signifies the best performance, while underlined text indicates the second-best performance (See Appendix B.1 for more details about the metrics).}
  \label{nsd results}
  \vspace{-0.3cm}
\end{table*}
\section{Experiment}
\subsection{Datasets}
Natural Scenes Dataset (NSD)\footnote{\url{https://naturalscenesdataset.org}}~\cite{nsd} is an extensive 7T fMRI dataset gathered from 8 subjects viewing images from the MSCOCO-2017 dataset, which contains images of complex natural scenes. Participants viewed three repetitions of 10,000 images with a 7-Tesla fMRI scanner over 30–40 sessions, with one session including 750 fMRI trials lasting for 1 hour. Details of the dataset information can be found in Appendix A. In this paper, we conducted 4 types of experiments, image retrieval and reconstruction in the next section, and text retrieval and brain correlation experiments in Appendix C.

\subsection{Implementation details}
All of our fine-tuning experiments were run for 150 epochs on two Tesla v100 32GB GPUs with a batch size of 10. The experimental parameter settings for 1-hour and 40-hour data are consistent. For fine-tuning experiments, the loss weight is set to $[\alpha_1,\alpha_2,\alpha_3,\alpha_4]=[0.5,1.0,1.5,0.5]$ and the rank $r$ is set to 8 for all LoRA blocks, including Skip-LoRAs. We use the AdamW~\cite{loshchilov2017adamw} for optimization, with a learning rate set to 3e-4, to which the OneCircle learning rate schedule~\cite{smith2019super} was set. During the training process, we use data augmentation from images and blurry images and replace the BiMixCo with SoftCLIP~\cite{mindseye2023} loss in one-third of the training phase. In the inference stage, we only generated a reconstructed image once and did not make multiple selections. The final high-level reconstructed image and low-level blurry image are simply weighted and averaged in a ratio of 3:1. In the inference stage of semantic correction, we use 80 category nouns from MSCOCO as correction texts and classify them in the form of text retrieval.
\vspace{-0.15cm}
\section{Results and Analysis}
\subsection{Image and Brain Retrieval}
Image retrieval refers to retrieving the image embeddings with the highest cosine similarity based on fMRI embeddings on the test set. In Table \ref{nsd results}, there is only a slight improvement in retrieval accuracy within 40 hours of data. This can be attributed to the fact that the retrieval accuracy is already close to the upper limit (only about 1\% short of reaching 100\%), which is affected by the noise of the fMRI dataset itself. However, when only 1 hour of data was available, MindTuner's retrieval accuracy was significantly higher than MineEye2, 4.1\% higher for image retrieval and 18.6\% higher for brain retrieval. As MindEye2 benefits from multi-subject pre-training and new-subject fine-tuning with linear heads, this suggests that the visual fingerprint we introduced significantly improves the performance of the model when fMRI data are scarce.
\vspace{-0.3cm}
\begin{figure}[!h]
  \centering
   \includegraphics[width=1.0\linewidth]{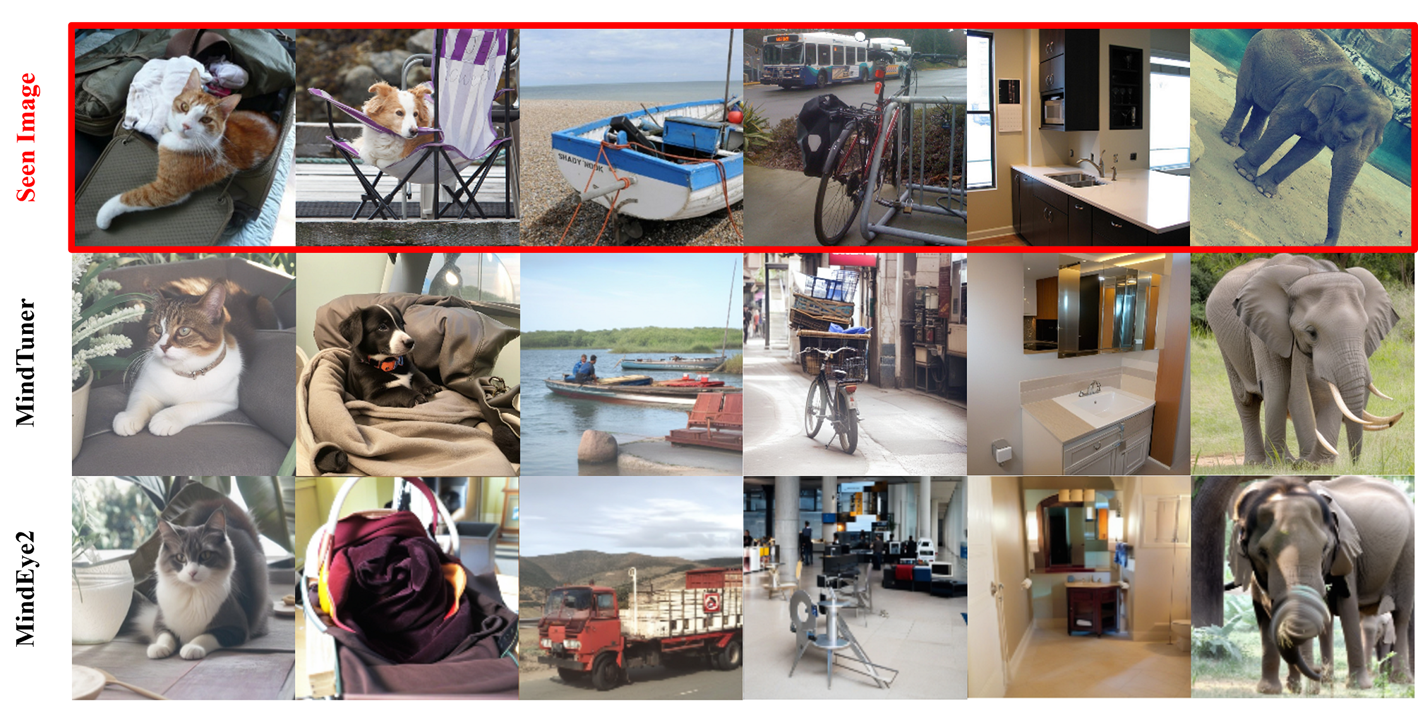}
   \vspace{-0.5cm}
   \caption{MindTuner vs MindEye2 reconstructions from fMRI brain activity with only 1 hour of data.}
   \vspace{-0.4cm}
   \label{nsd comparision 1 session}
\end{figure}
\begin{figure}[!h]
  \centering
   \includegraphics[width=1.0\linewidth]{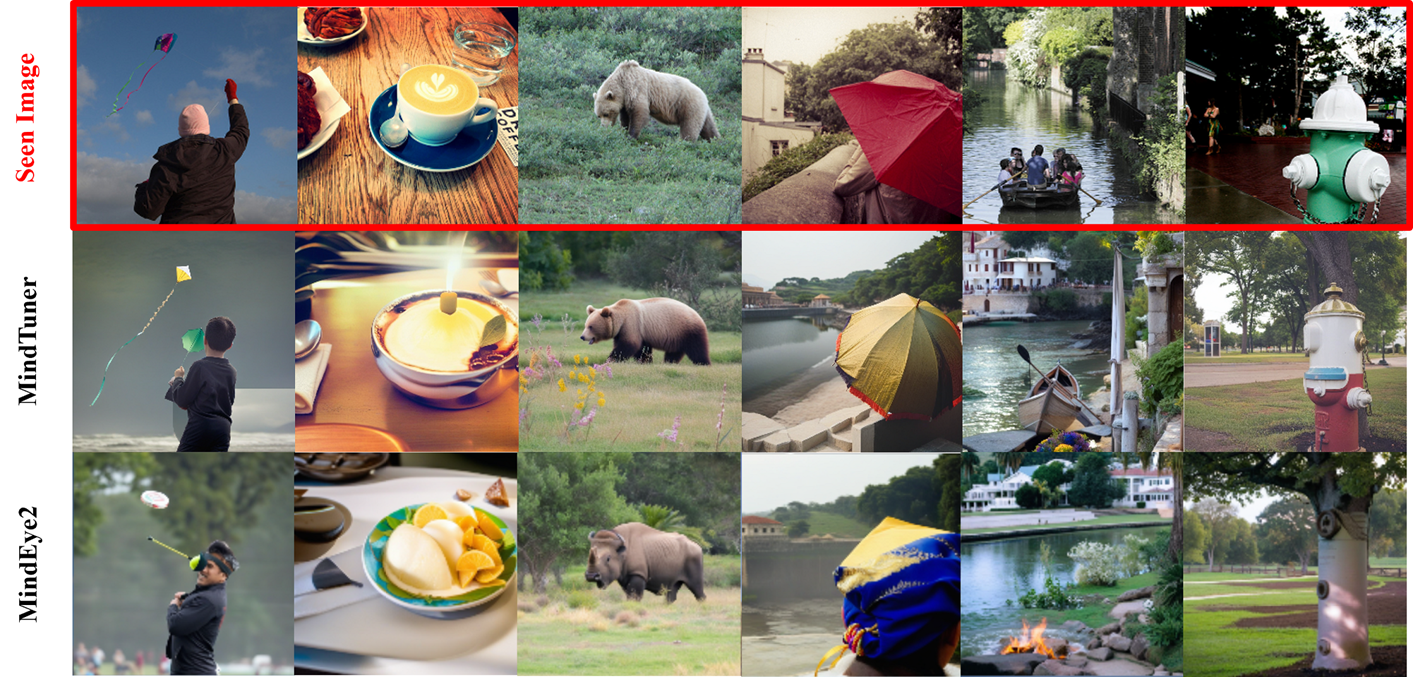}
   \caption{MindTuner vs MindEye2 reconstructions from fMRI brain activity with 40 hours of data.}
   \vspace{-0.6cm}
   \label{nsd comparision 40 sessions}
\end{figure}

\vspace{-0.2cm}
\begin{table*}[t]
  \scalebox{0.83}{
  \centering
  \begin{tabular}{@{}cccccccccccc@{}}
    \toprule
    \multirow{2}{*}{Method} &Trainable&\multicolumn{4}{c}{Low-Level}& \multicolumn{4}{c}{High-Level}& \multicolumn{2}{c}{Retrieval}\\
    \cmidrule(lr){3-6} \cmidrule(lr){7-10} \cmidrule(lr){11-12}
     &Parameters&PixCorr$\uparrow$&SSIM$\uparrow$&Alex(2)$\uparrow$&Alex(5)$\uparrow$&Incep$\uparrow$&CLIP$\uparrow$&Eff$\downarrow$&SwAV$\downarrow$&Image$\uparrow$&Brain$\uparrow$\\

    \midrule
     MindEye2&64.4M(-)&0.235&\textbf{0.428}&88.0\%&93.3\%&83.6\%&80.8\%&0.798&0.459&94.0\%&77.6\%\\
     with Adaptive Projector&66.5M(\textbf{+2.1M})&0.233&0.426&87.8\%&93.0\%&84.0\%&81.2\%&0.794&0.454&93.8\%&77.3\%\\
     with only LoRAs&74.6M(\textbf{+9.4M})&0.261&0.427&90.3\%&94.2\%&84.8\%&84.0\%&0.784&0.441&93.7\%&85.6\%\\
     LoRAs+Skip-LoRAs&74.6M(\textbf{+10.2M})&\textbf{0.264}&0.427&\textbf{90.8\%}&94.8\%&85.1\%&84.2\%&0.780&0.437&\textbf{94.5\%}&\textbf{87.7\%}\\
     MindTuner(non-linear)&76.7M(\textbf{+12.3M})&0.191&0.383&82.7\%&88.0\%&77.3\%&76.0\%&0.848&0.492&27.4\%&56.7\%\\
     \rowcolor{lightblue}
     MindTuner&76.7M(\textbf{+12.3M})&0.262&0.422&90.6\%&\textbf{94.9\%}&\textbf{85.8\%}&\textbf{84.6\%}&\textbf{0.774}&\textbf{0.433}&\underline{94.2\%}&\underline{87.4\%}\\
     \bottomrule
  \end{tabular}
  }
  \vspace{-0.15cm}
  \caption{Ablations on the modules of MindTuner. Note that all model parameters for MindEye2 are 2.2B, and we only reported the trainable parameters required for new-subject fine-tuning. MinTuner(non-linear) denotes adding an activation function to the ridge regression head.}
  \vspace{-0.3cm}
  \label{ablations}
\end{table*}

\begin{table*}[t]
  \scalebox{0.85}{
  \centering
  \begin{tabular}{@{}ccccccccccccc@{}}
    \toprule
    \multirow{2}{*}{$rank$} &Trainable&Skip-LoRAs&\multicolumn{4}{c}{Low-Level}& \multicolumn{4}{c}{High-Level}& \multicolumn{2}{c}{Retrieval}\\
    \cmidrule(lr){4-7} \cmidrule(lr){8-11} \cmidrule(lr){12-13}
     &Parameters&Parameters&PixCorr$\uparrow$&SSIM$\uparrow$&Alex(2)$\uparrow$&Alex(5)$\uparrow$&Incep$\uparrow$&CLIP$\uparrow$&Eff$\downarrow$&SwAV$\downarrow$&Image$\uparrow$&Brain$\uparrow$\\
     
     \midrule
     $4$&71.6M(+\textbf{7.2M})&0.4M&0.263&0.422&90.4\%&94.9\%&84.8\%&84.6\%&0.778&0.436&94.4\%&87.9\%\\
     \rowcolor{lightblue}
     $8$&76.7M(\textbf{+12.3M})&0.8M&0.262&0.422&90.6\%&94.9\%&85.8\%&84.6\%&0.774&0.433&94.2\%&87.4\%\\
     $16$&86.7M(\textbf{+22.3M})&1.6M&0.262&0.422&90.5\%&95.1\%&85.2\%&85.2\%&0.776&0.434&93.7\%&87.2\%\\
     \bottomrule
  \end{tabular}
  }
  \vspace{-0.15cm}
  \caption{Ablations on the rank $r$ of MindTuner.}
  \vspace{-0.6cm}
  \label{ablations_r}
\end{table*}

\subsection{Image Reconstruction}
Image reconstruction aims to restore the original image as seen by the subjects, and two levels of evaluation metrics assess the quality of the reconstructions. Previous research has improved the performance of these above metrics in single-subject models by various means. The cross-subject task discussed in this paper tests the ability of the model to exploit the commonalities of multi-subject and to migrate new subjects, ultimately realizing the goal of using less fMRI data for few-shot learning. Here, we report the reconstruction results of MindTuner using 1-hour and 40-hour training data. Table \ref{nsd results} reflects the quantized performance comparisons, under two different training data sizes. It can be seen that when using the full NSD dataset, MindTuner achieves better performance on high-level metrics; when only 1 hour of data is available for training, MindTuner outperforms MindEye2 on all metrics. The visualization results of 1-hour and 40-hour reconstructions can be seen in Figure \ref{nsd comparision 1 session} and Figure \ref{nsd comparision 40 sessions}. It can be seen that the quality of the reconstructed images at 40 hours is significantly higher than at 1 hour as the training data increases. Meanwhile, our MindTuner is better than MindEye2 in both semantic completeness and category accuracy, demonstrating the superiority of the overall model. Our visualization excludes the other three methods in Table \ref{nsd results} because the image semantics generated by these three methods are very unclear. Interestingly, we also found that images generated directly from SDXL unCLIP outperformed MindTuner's corrected images at a high level, albeit visibly distorted. Further results are in Appendix D.1.

\section{Ablations}
In this section, we explored the effectiveness of each component of our method through ablation experiments. All results were pre-trained on Subjects 2-8, and fine-tuning was performed on Subject 1 using 1 hour of data.

\textbf{MindTuner's modules.} 
Further experiments were conducted to assess the efficacy of each module. As can be observed in Table \ref{ablations}, the incorporation of LoRAs markedly enhances the model's capacity. The incorporation of Skip-LoRAs resulted in a convergence of all image reconstruction metrics towards the performance of the complete model, with the retrieval accuracy even exceeding that of the complete model. This can be attributed to the balance of additional image-to-text aligned losses, which has the effect of slightly reducing the performance of the retrieval submodule in the complete model. Furthermore, the incorporation of an adaptive projector has enhanced the performance of high-level reconstruction, and additional visualizations in Figure \ref{class correction} have also demonstrated the efficacy of semantic correction. However, this approach resulted in the compromise of certain low-level effects. In comparison to the 2.2B multi-subject pre-trained main model, MindTuner resulted in a mere 12.3M increase in parameters (0.56\%), yet achieved superior outcomes. Furthermore, an activation function was added to the ridge head to make the entire alignment head non-linear. However, this resulted in a serious overfitting phenomenon, with all metrics exhibiting a significant decline, which indicates that the Skip-LoRAs in MindTuner offer a more suitable alternative for non-linear fMRI alignment of diverse subjects.

\textbf{The rank $r$ of Skip-LoRAs and LoRAs.} Further experiments were conducted on the most critical variable in the LoRAs, namely the rank $r$. As can be observed in Table \ref{ablations_r}, when the rank $r=4/16$, the overall ability of the model does not show significant changes. This finding is consistent with the results presented in the LoRA paper~\cite{hu2021lora}, which highlighted that when r is within the range of 2 to 16, the fine-tuning effect remains unchanged. Furthermore, no substantial evidence of overfitting was identified. This can be attributed to the fact that the parameters in Skip-LoRAs have remained relatively modest, below 2M.

\vspace{-0.2cm}
\section{Neuroscience interpretability}
\begin{figure}[h]
  \centering
   \includegraphics[width=1.0\linewidth]{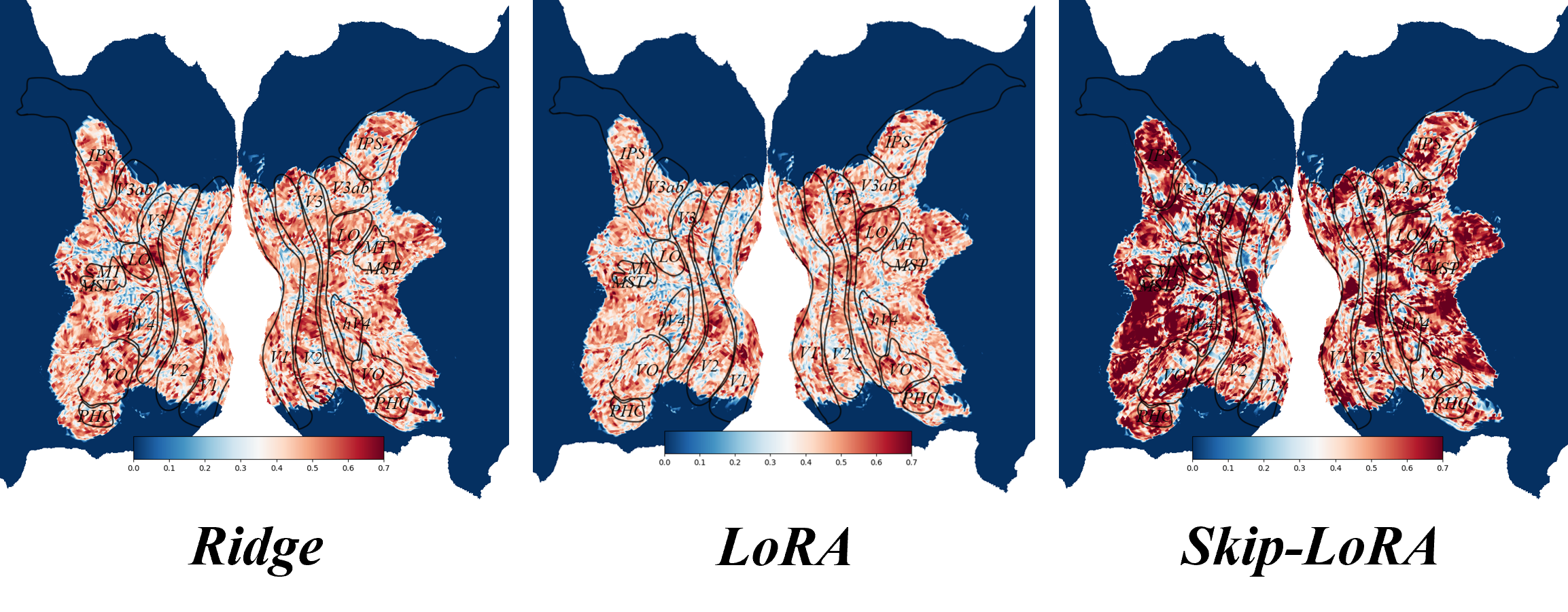}
   \vspace{-0.7cm}
   \caption{The importance of different ROIs of subject 1 with fitted weights of the first layer. The weights of each module are averaged and normalized between 0 and 1.}
   \vspace{-0.4cm}
   \label{neural}
\end{figure}

To investigate the interpretability of MindTuner in neuroscience, we conduct experiments to explore where the subject's non-linear relationship comes from in the visual cortex. We utilize pycortex~\cite{gao2015pycortex} to project the weights of each voxel in the first layer of Ridge regression, LoRA, or Skip-LoRA onto the corresponding 2D flat map of the NSD dataset. The results are presented in Figure \ref{neural}. For linear heads, the importance of visual cortical voxels is similar for both ridge and LoRA. However, for Skip-LoRA, a small portion of the early visual cortex and more advanced visual cortex are valued more. This indicates that at the fMRI level, a greater concentration of non-linear relationships is found within the higher visual cortex. It also demonstrates that the design of Skip-LoRAs is capable of capturing non-linear relationships in fMRI data that are not discernible with standard LoRAs. More visualization results, as well as brain region ROI templates, can be found in Appendix E.
\vspace{-0.3cm}

\section{Conclusion}
In this paper, we propose MindTuner, a new cross-subject decoding method. We introduced the phenomenon of visual fingerprint in the human visual system and utilized the combination of Skip-LoRAs and LoRAs to learn each subject's visual fingerprint. Meanwhile, we innovatively propose a method for enhancing reconstruction by indirectly connecting fMRI with text in visual decoding tasks. Experimental results have shown that we have achieved better performance on multiple evaluation metrics at a relatively small parameter cost, especially when the fMRI data is insufficient. Our work has relaxed the conditions for fMRI acquisition, helping to achieve a universal brain decoding model in the future.

\section*{Acknowledgements}
We have included more experimental results and analyses in the appendix, which you can access through this \textit{arXiv link}\footnote{\url{https://arxiv.org/abs/2404.12630}}.

\noindent This research is supported by the National Key Research and Development Program of China (No.~2022YFB3104700), the National Natural Science Foundation of China (No.~61976158, No.~62376198), Shanghai Baiyulan Pujiang Project (No.~08002360429). 

\bibliography{aaai25}

\newpage
\section{Additional Details}
\subsection{Evaluation Metrics}
\textit{\textbf{Retrieval}}: Image retrieval refers to retrieving the image embeddings with the highest cosine similarity based on voxel embeddings on the test set (chance=0.3\%). If a paired image embedding is retrieved, the retrieval is considered
correct. Brain retrieval is the opposite process mentioned above.

\noindent\textit{\textbf{PixCorr}}: pixel-wise correlation between ground truth and reconstructions; 

\noindent\textit{\textbf{SSIM}}: structural similarity index metric~\cite{ssim} between ground truth and reconstructions;

\noindent\textit{\textbf{Eff}}~\cite{eff}and \textit{\textbf{Swav}}~\cite{swav} refer to average correlation distance with EfficientNet-B1 and SwAV-ResNet50.

\noindent\textit{\textbf{Alex(2), Alex(5), Incep, CLIP}}: all these metrics refer to two-way identification (chance = 50\%) using different models. The two-way comparisons were performed with AlexNet where Alex(2) denotes the second layer, Alex(5) denotes the fifth layer, InceptionV3 with the last pooling layer, and CLIP with the final layer of ViT-L/14. Two-way identification refers to percent correct across comparisons gauging if the original image embedding is more similar to its paired voxel embedding or a randomly selected voxel embedding. We followed the same image preprocessing and two-way identification steps as ~\cite{brain-diffuser, mindseye2023, scotti2024mindeye2}. For each test sample, performance was averaged across all possible pairwise comparisons using the other 999 reconstructions to ensure no bias from random sample selection. This yielded 1,000 averaged percent correct outputs.

\begin{figure}[ht]
  \centering
   \includegraphics[width=0.5\linewidth]{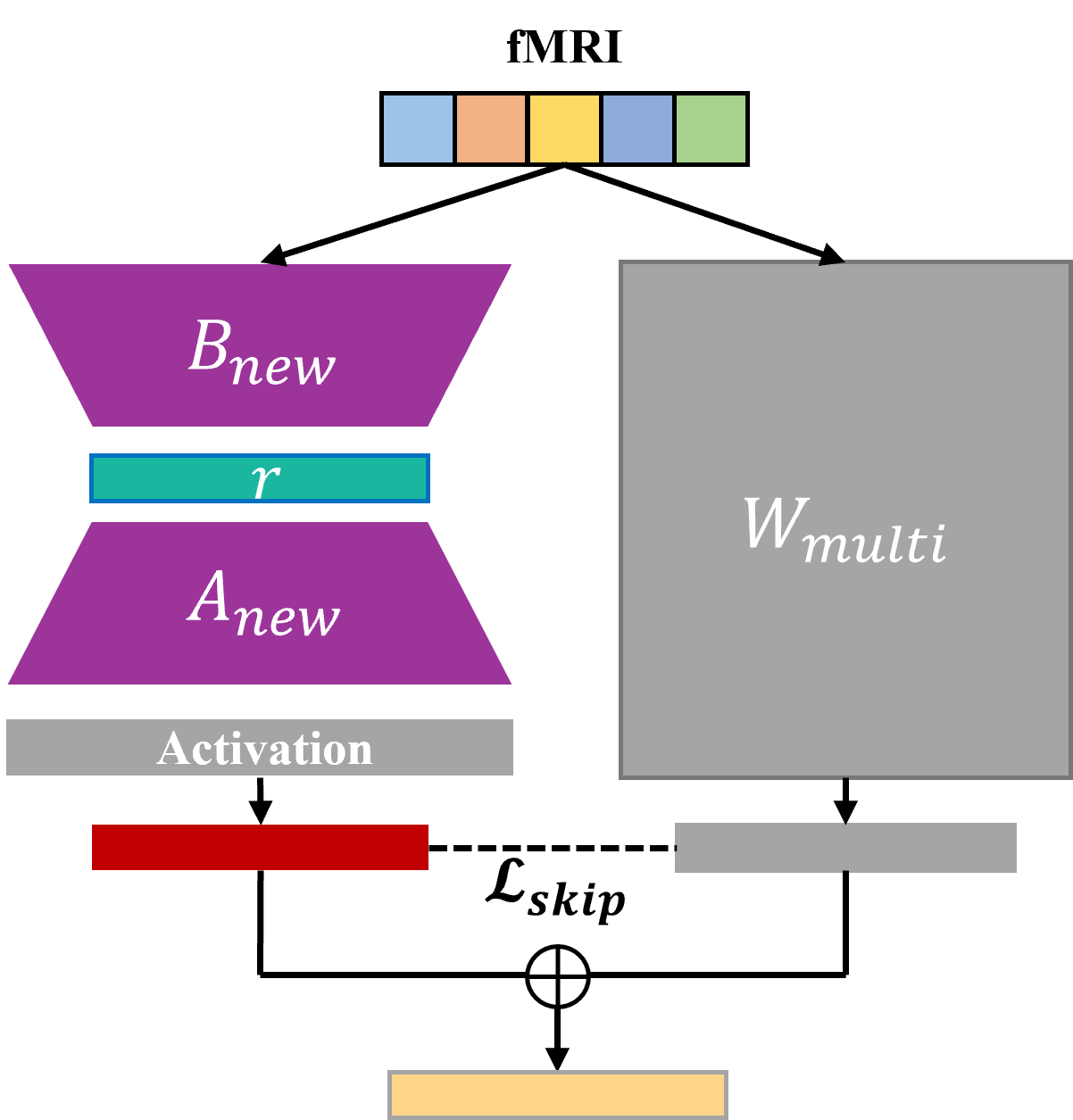}
   \caption{Design of Skip-LoRAs. $W_{multi}$ denotes shared weight during multi-subject pre-training. A Skip-LoRA block consists of a LoRA block with skip-connection, activation function, and non-linear loss $\mathcal{L}_{skip}$}
   \label{skip-loras}
\end{figure}

\subsection{Skip-LoRAs}
As described in the main text, we designed Skip-LoRAs to learn better visual fingerprint between subjects, especially the non-linear part. The structure of Skip-LoRAs is shown in Figure \ref{skip-loras}. The pre-trained shared model of the MLP Backbone includes an alignment layer that aligns voxels to 4096-dim, four 4096-dim residual blocks, and a layer that maps to image tokens. We used Skip-LoRAs to connect fMRI to these layers, allowing the initial fMRI differences to affect the entire network. $\mathcal{L}_{skip}$ was also used for these connections, except for the last mapping layer of image tokens, as the non-linear constraints and retrieved cosine similarity-based contrastive loss conflicted with each other.

\section{Additional Results}
\subsection{Adapative Porjector pre-trained on MSCOCO}
Previous CLIP-related articles have demonstrated that ViT's projection layer can take both CLS tokens and the result of the global average pooling of all tokens as inputs. We perform a global average pooling of all 256 brain tokens, which finally are mapped to the same 1280 dimensions as the CLS token of the text. The PyTorch code used to train the projector is depicted below:
\begin{lstlisting}[caption={}]
class Adaptive_Projector(torch.nn.Module):
    def __init__(self):
        super().__init__()
        self.proj = nn.Parameter(
                    torch.randn(1664, 1280))
    def forward(self, x):
        x = torch.mean(x, dim = 1)
        x = x @ self.proj
        return x
\end{lstlisting}
We trained this projection layer on the whole COCO 2017 dataset including 73k images and 5 texts for each image. We randomly selected 900 image-text pairs as the validation set, and used a retrieval pool of 300 for validation (a similar setup to our main experiment). Instead of training from scratch, we used the projection layer of image CLS token from the open-source OpenCLIP\footnote{\url{https://huggingface.co/laion/CLIP-ViT-bigG-14-laion2B-39B-b160k}} to start fine-tuning for 20 epochs, with the learning rate being fixed at 3e-4 and the batch size at 512. The loss is the standard CLIP image-text contrastive loss. The accuracy of validation set retrieval before and after fine-tuning is as follows:
\begin{table}[h]
    \centering
    \begin{tabular}{c|c|c}
    \toprule
         Method&Image2Text&Text2Image  \\
    \midrule
         Before fine-tuning&81.3\%&78.7\% \\
         After fine-tuning&\textbf{95.2\%}&\textbf{95.2\%} \\
    \bottomrule
    \end{tabular}
    \caption{Fine-tuning results on the COCO 2017.}
    \label{COCO fine-tuning}
\end{table}

\begin{table*}[t]
  \centering
  \begin{tabular}{@{}cccccccccc@{}}
    \toprule
    \multirow{2}{*}{Method}&\multirow{2}{*}{Data Size}&\multicolumn{4}{c}{Low-Level}& \multicolumn{4}{c}{High-Level}\\
    \cmidrule(lr){3-6} \cmidrule(lr){7-10}
     &&PixCorr$\uparrow$&SSIM$\uparrow$&Alex(2)$\uparrow$&Alex(5)$\uparrow$&Incep$\uparrow$&CLIP$\uparrow$&Eff$\downarrow$&SwAV$\downarrow$\\
     
     \midrule
     MindTuner(Uncorrected)&40 hours&0.280&0.327&95.4\%&99.2\%&96.4\%&94.5\%&0.621&0.341\\
     \rowcolor{lightblue}
     MindTuner&40 hours&0.322&0.421&95.8\%&98.8\%&95.6\%&93.8\%&0.612&0.340\\
     \midrule
     MindTuner(Uncorrected)&1 hour&0.208&0.378&87.4\%&94.0\%&85.5\%&83.8\%&0.781&0.439\\
     \rowcolor{lightblue}
     MindTuner&1 hour&0.224&0.420&87.8\%&93.6\%&84.8\%&83.5\%&0.780&0.440\\
     \bottomrule
  \end{tabular}
  \vspace{0.2cm}
  \caption{Quantitative effects on semantic correction.}
  \vspace{-0.3cm}
  \label{semantic correction}
\end{table*}
\subsection{Semantic Correction Influence}
As shown in Figure \ref{semantic correction}, we observe that semantic correction, while making the semantics of the reconstructed images clearer and more accurate, negatively affects the evaluation metrics, especially high-level. This is similar to MindEye2's results, suggesting that manual correction of the image, either by MindEye2's refinement or MindTuner's correction, can have an adverse effect on the original fMRI representation. And as shown in Table 1, MindTuner mitigates this negative effect to some extent, compared to MindEye2.
\subsection{Additional Brain Correlation Reults}
The Brain Correlation results with 1-hour data are as follows. In conjunction with Table 3, it can be seen that MindTuner greatly improves the brain correlation of the reconstructed images, especially when the data is scarce. It suggests that the learning of subjects' visual fingerprints preserves their own properties to some extent.

\begin{table}[h]
  \centering
  \scalebox{0.8}{
  \begin{tabular}{@{}cccc@{}}
    \toprule
    Brain Region&MindTuner&MindEye2~\cite{scotti2024mindeye2}\\
    \midrule
     Visual cortex$\uparrow$&\textbf{0.372}&0.348\\
     V1$\uparrow$&\textbf{0.345}&0.309\\
     V2$\uparrow$&\textbf{0.348}&0.314\\
     V3$\uparrow$&\textbf{0.347}&0.315\\
     V4$\uparrow$&\textbf{0.329}&0.300\\
     Higher vis.$\uparrow$&\textbf{0.370}&0.351\\
     \bottomrule
  \end{tabular}
  }
  \caption{The brain correlation scores computed across various brain regions with 1 hour data.}
  \vspace{-3mm}
  \label{brain correlation}
\end{table}

\begin{table}[h]
  \scalebox{0.75}{
  \centering
  \begin{tabular}{@{}c|c|c|c|c|c@{}}
    \toprule
    \multicolumn{2}{c}{\textbf{1-hour data}}|&Subject 1&Subject 2&Subject 5&Subject 7\\
    \midrule
    \multirow{4}{*}{Low-Level}&PixCorr$\uparrow$&\textbf{0.262}&0.225&0.208&0.202\\
    &SSIM$\uparrow$&0.422&\textbf{0.425}&0.415&0.417\\
    &Alex(2)$\uparrow$&\textbf{90.6\%}&89.1\%&86.8\%&84.5\%\\
    &Alex(5)$\uparrow$&94.9\%&\textbf{95.1\%}&93.7\%&90.8\%\\
    \midrule
    \multirow{4}{*}{High-Level}&Incep$\uparrow$&85.8\%&84.8\%&\textbf{87.7\%}&80.7\%\\
    &CLIP$\uparrow$&84.6\%&83.7\%&\textbf{85.9\%}&79.6\%\\
    &Eff$\downarrow$&0.774&0.781&\textbf{0.750}&0.817\\
    &SwAV$\downarrow$&0.433&0.440&\textbf{0.422}&0.465\\
    \midrule
    \multirow{2}{*}{Retrieval}&Image$\uparrow$&\textbf{94.2\%}&93.7\%&72.2\%&71.9\%\\
    &Brain$\uparrow$&\textbf{87.4\%}&82.8\%&68.1\%&65.5\%\\
    \midrule
     \multirow{4}{*}{Brain}&Visual cortex$\uparrow$&0.359&0.376&\textbf{0.426}&0.328\\
     &V1$\uparrow$&0.336&0.342&\textbf{0.366}&0.335\\
     \multirow{3}{*}{Correlation}&V2$\uparrow$&0.354&0.326&\textbf{0.373}&0.339\\
     &V3$\uparrow$&\textbf{0.355}&0.348&0.354&0.329\\
     &V4$\uparrow$&0.327&\textbf{0.366}&0.328&0.294\\
     &Higher vis.$\uparrow$&0.358&0.379&\textbf{0.432}&0.311\\
     \bottomrule
  \end{tabular}
  }
  \caption{Specific subject quantitative results with 1-hour training data.}
  \label{specific 1 hour}
\end{table}

\begin{table}[h]
  \scalebox{0.75}{
  \centering
  \begin{tabular}{@{}c|c|c|c|c|c @{}}
    \toprule
    \multicolumn{2}{c}{\textbf{40-hours data}}|&Subject 1&Subject 2&Subject 5&Subject 7\\
    \midrule
    \multirow{4}{*}{Low-Level}&PixCorr$\uparrow$&\textbf{0.371}&0.331&0.298&0.288\\
    &SSIM$\uparrow$&\textbf{0.428}&0.421&0.422&0.414\\
    &Alex(2)$\uparrow$&\textbf{97.7\%}&96.8\%&94.7\%&94.1\%\\
    &Alex(5)$\uparrow$&\textbf{99.3\%}&99.0\%&98.7\%&98.1\%\\
    \midrule
    \multirow{4}{*}{High-Level}&Incep$\uparrow$&96.4\%&95.1\%&\textbf{96.6\%}&94.1\%\\
    &CLIP$\uparrow$&94.3\%&92.7\%&\textbf{94.7\%}&93.3\%\\
    &Eff$\downarrow$&0.601&0.620&\textbf{0.592}&0.636\\
    &SwAV$\downarrow$&\textbf{0.331}&0.342&0.333&0.355\\
    \midrule
    \multirow{2}{*}{Retrieval}&Image$\uparrow$&\textbf{100\%}&99.9\%&98.4\%&97.2\%\\
    &Brain$\uparrow$&\textbf{99.9\%}&99.8\%&96.8\%&96.6\%\\
    \midrule
     \multirow{4}{*}{Brain}&Visual cortex$\uparrow$&0.377&0.394&\textbf{0.416}&0.320\\
     &V1$\uparrow$&0.387&\textbf{0.400}&0.357&0.323\\
     \multirow{3}{*}{Correlation}&V2$\uparrow$&\textbf{0.381}&0.359&0.361&0.317\\
     &V3$\uparrow$&0.366&\textbf{0.373}&0.341&0.302\\
     &V4$\uparrow$&0.340&\textbf{0.376}&0.324&0.278\\
     &Higher vis.$\uparrow$&0.363&0.387&\textbf{0.426}&0.311\\
     \bottomrule
  \end{tabular}
  }
  \caption{Subject's specific quantitative results with 40-hours training data.}
  \vspace{-3mm}
  \label{specific 40 hours}
\end{table}
\subsection{Subject's Specific Results}
Tables \ref{specific 1 hour} and \ref{specific 40 hours} show more exhaustive evaluation metrics computed for every subject individually using 40-hours and 1-hour of fine-tuning data respectively. It can be seen that the performance of different subjects is relatively similar regardless of the amount of data, e.g., Subject 1 performs better at the low level and retrieval, while Subject 5 performs better at the high level.
\subsection{Subject's Specific Visualizations}
\begin{figure*}[h]
  \centering
   \includegraphics[width=0.96\linewidth]{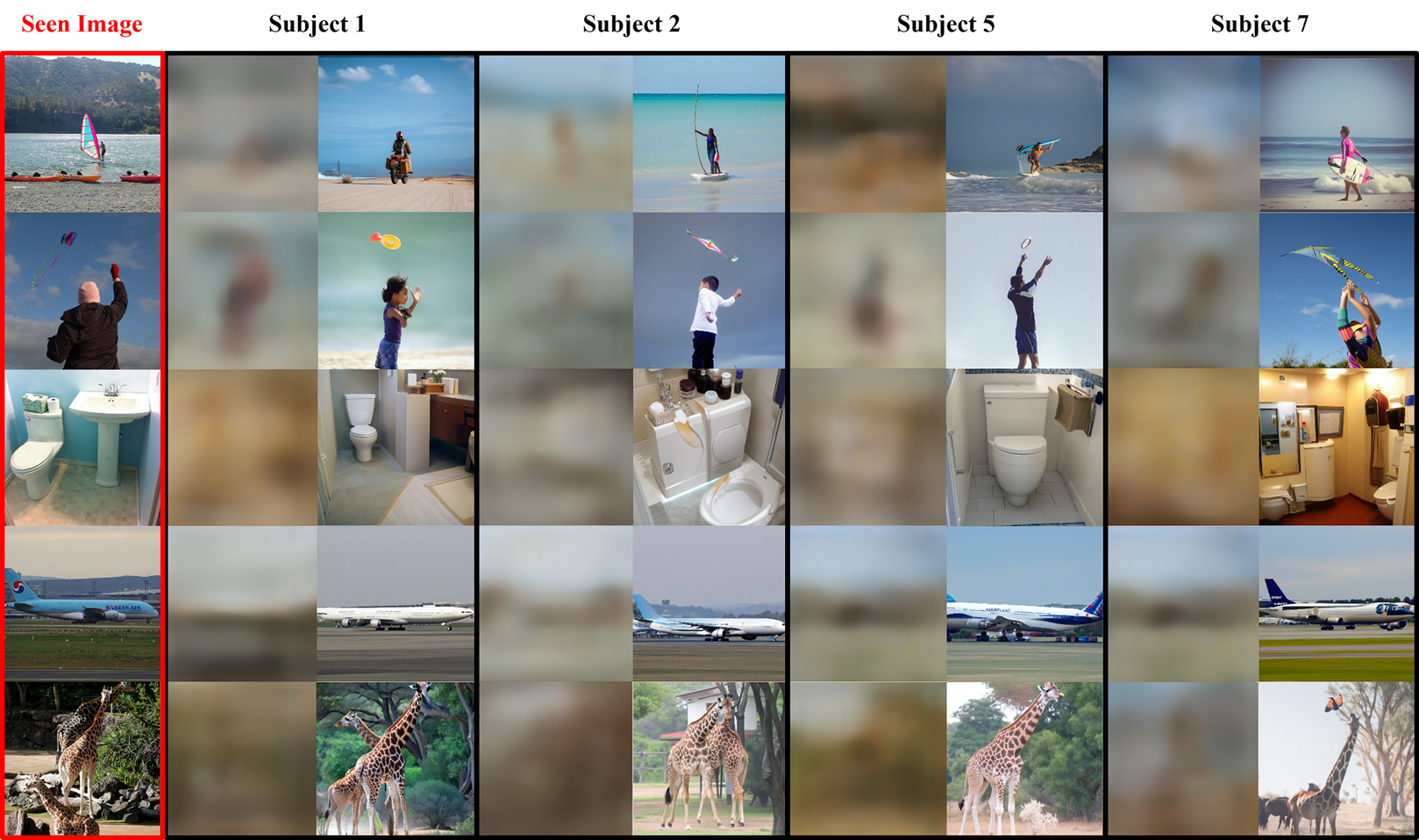}
   \caption{Subject's specific visualizations of MindTuner with 1-hour training data.}
   \vspace{-1.5mm}
   \label{1 hour results}
\end{figure*}

\begin{figure*}[h]
  \centering
   \includegraphics[width=0.96\linewidth]{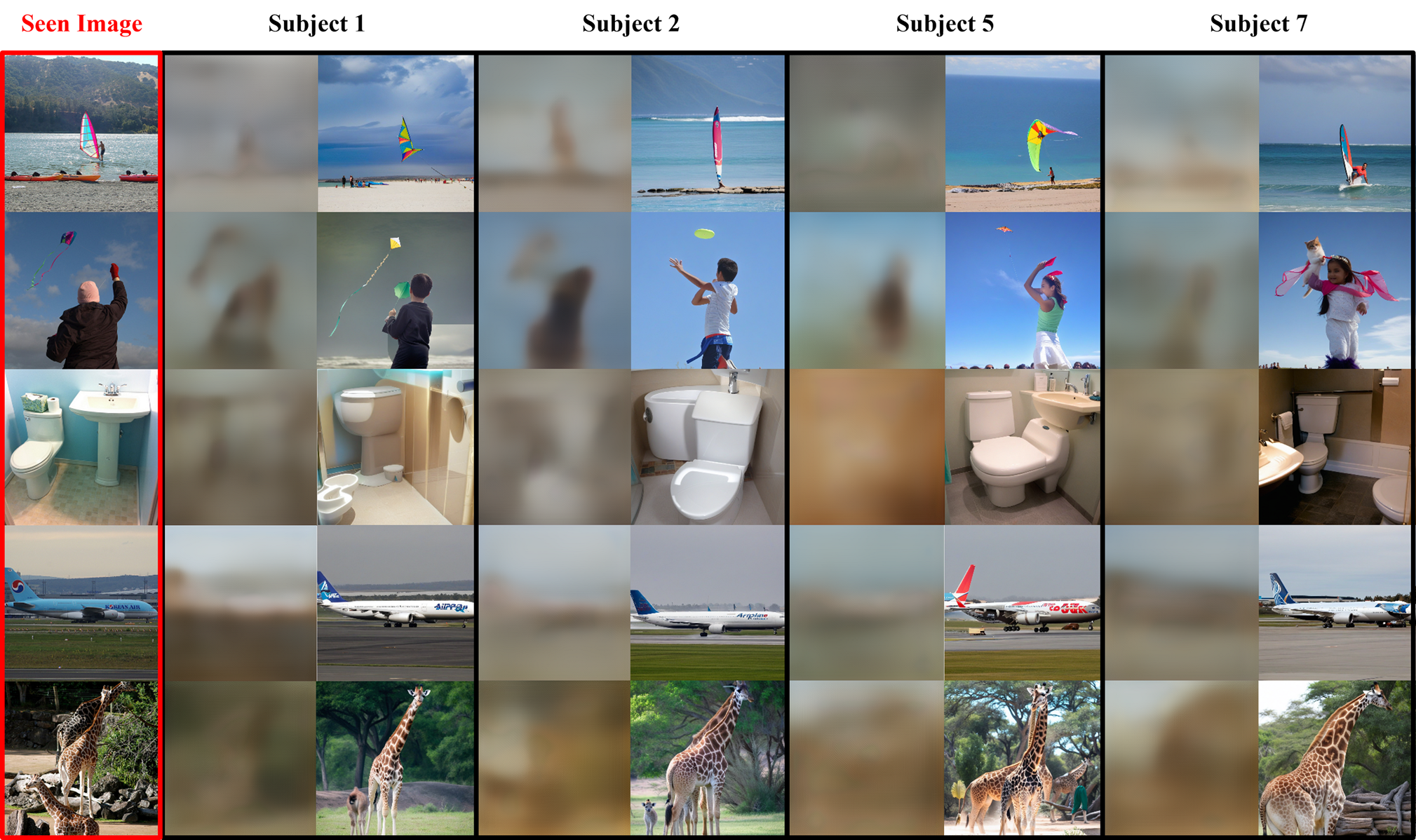}
   \caption{Subject's specific visualizations of MindTuner with 40-hours training data.}
   \label{40 hours results}
\end{figure*}

We visualize in Figure \ref{1 hour results} and Figure \ref{40 hours results} the specific reconstruction results for different subjects, using either 1-hour or 40-hours data. For different subjects, MindTuner's reconstructed images were accurate in capturing semantics, but the details varied. This may be attributed to the fact that the CLIP representation space is more semantic than detailed texture. In addition to this, as the training data increases, the blurry images become more similar to the ground truth images, thus giving a boost to the low-level metrics, as shown in Tables \ref{1 hour results} and \ref{40 hours results}. Unfortunately, the color distribution of blurry images and reconstructed images is often difficult to control, e.g., generating airplanes of various colors. We have tried to add a sub-module to align the fMRI to a $16\times16$ spatial palette, but the spatial palettes obtained are only slightly better than the blurry images. Moreover, the T2I Adapter generation model based on a spatial palette is difficult to control, which often makes the reconstruction results even worse. Thus, the relationship between the visual system and color needs to be further explored.

\subsection{COCO Category}
There are 80 categories of object categorization in MSCOCO, and the category text was used by us for semantic correction in MindTuner, below are the 80 categories:

\noindent\textit{'person, bicycle, car, motorcycle, airplane, bus, train, truck, boat, traffic light, fire hydrant, stop sign, parking meter, bench, bird, cat, dog, horse, sheep, cow, elephant, bear, zebra, giraffe, backpack, umbrella, handbag, tie, suitcase, frisbee, skis, snowboard, sports ball, kite, baseball bat, baseball glove, skateboard, surfboard, tennis racket, bottle, wine glass, cup, fork, knife, spoon, bowl, banana, apple, sandwich, orange, broccoli, carrot, hot dog, pizza, donut, cake, chair, couch, potted plant, bed, dining table, toilet, tv, laptop, mouse, remote, keyboard, cell phone, microwave, oven, toaster, sink, refrigerator, book, clock, vase, scissors, teddy bear, hair drier, toothbrush'}
\subsection{Failure Results}
\begin{figure}[h]
  \centering
   \includegraphics[width=0.8\linewidth]{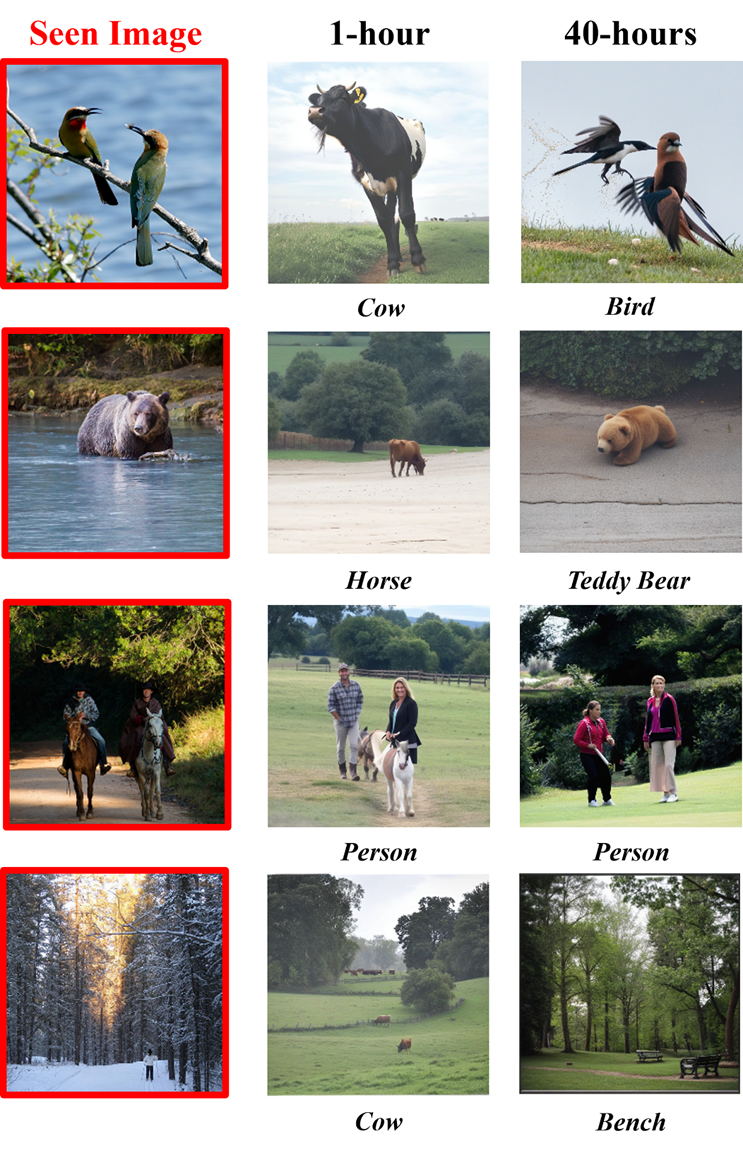}
   \vspace{-1.5mm}
   \caption{Failure Reconstructions.}
   \vspace{-4mm}
   \label{failure results}
\end{figure}

In this section, we visualize some of the failed reconstructed images, and we mainly identify semantically incorrect images as failed reconstructions because if the semantics are incorrect, it is even less important to talk about other details. All results shown in Figures \ref{failure results} are from Subject 1.

The first reconstruction failures are purely due to insufficient training data, e.g., the bear in the first line and the bird in the second line have bad semantics when only 1 hour of training data is available, but the semantics are progressively more accurate over 40 hours. The second type of reconstruction failure is caused by similar categories resulting in insufficient semantic accuracy, e.g., bears and teddy bears, cakes and pizzas, etc. The third type of reconstruction failure is due to the inability of category nouns to describe overly complex scenarios, especially when a person is present. Since MSCOCO's category noun is only 'person', it often appears that the person determines the categorization result while ignoring other objects, and the gender of the person is often not accurately generated. This is because MindTuner only uses a single category noun, so more complex Pivot designs can be discussed in the future, whether for text generation or multi-classification.

\section{Additional Neuroscience Visualizations}
\begin{figure}[h]
  \centering
   \includegraphics[width=1.0\linewidth]{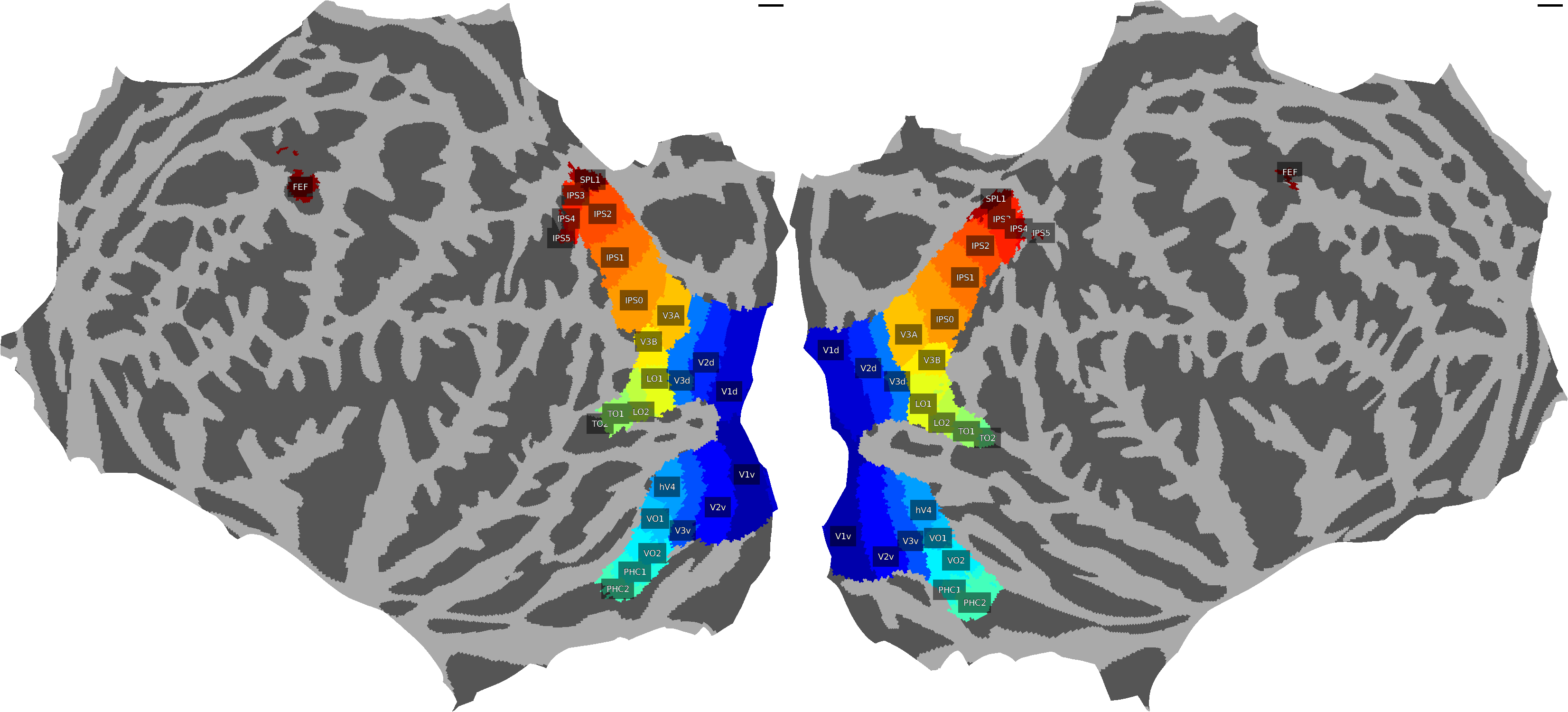}
   \caption{ROI reference map of visual cortex with template fsaverage.}
   \vspace{-1.5mm}
   \label{fsaverage}
\end{figure}

\begin{figure*}[h]
  \centering
   \includegraphics[width=0.8\linewidth]{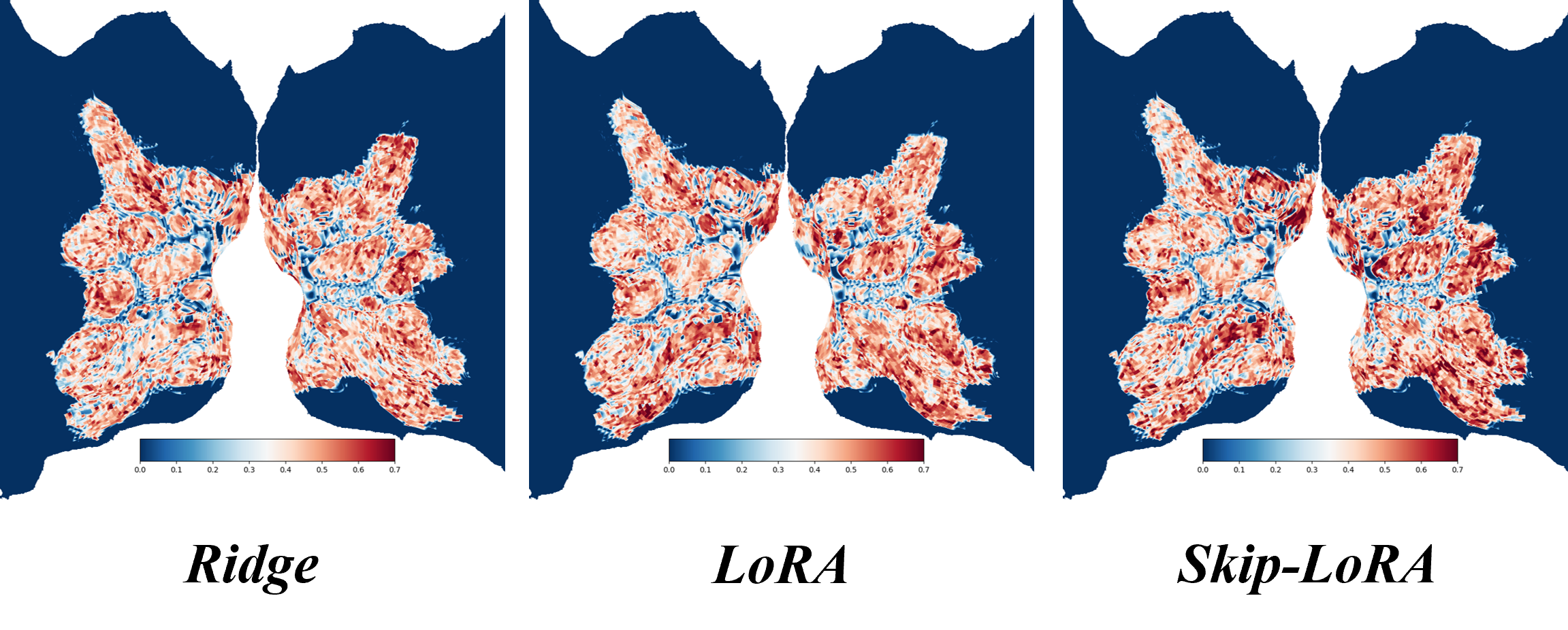}
   \caption{The importance of different ROIs of Subject 2 with
fitted weights of the first layer.}
   \label{subj02}
\end{figure*}

\begin{figure*}[h]
  \centering
   \includegraphics[width=0.8\linewidth]{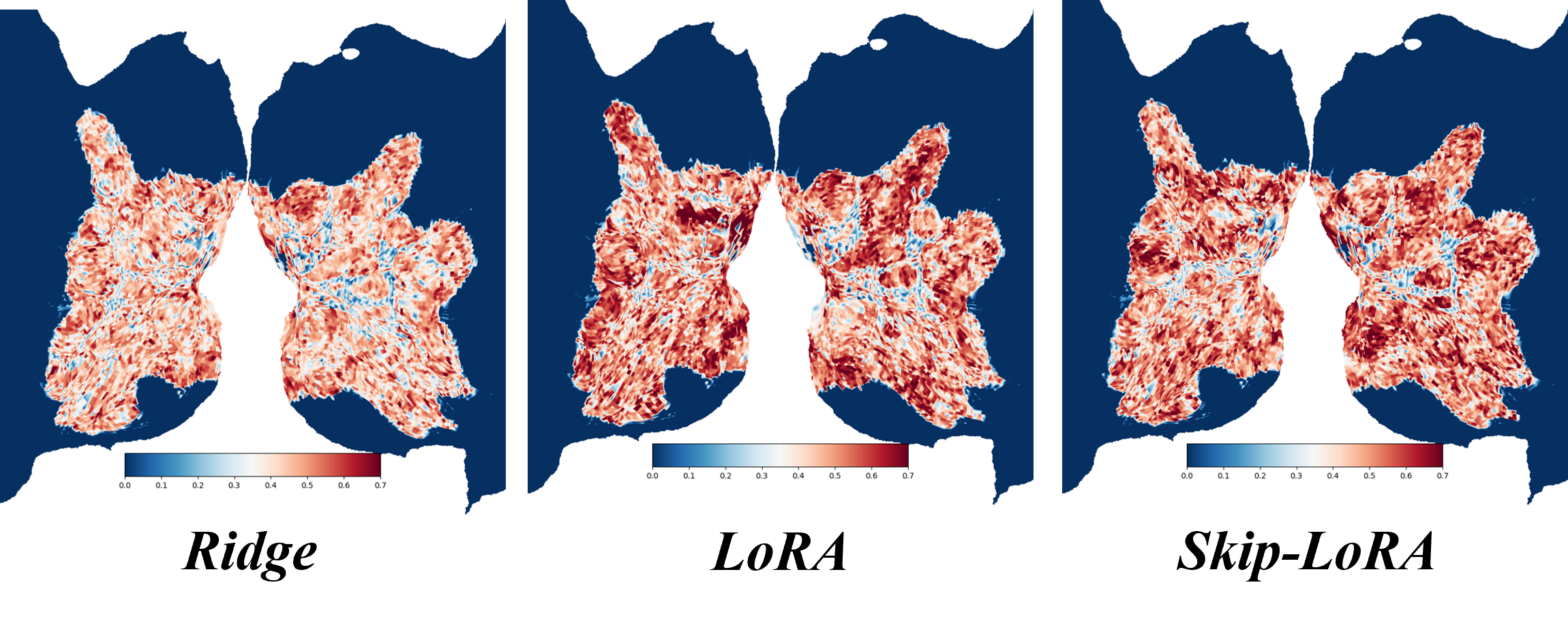}
   \caption{The importance of different ROIs of Subject 5 with
fitted weights of the first layer.}
   \label{subj05}
\end{figure*}

\begin{figure*}[h]
  \centering
   \includegraphics[width=0.8\linewidth]{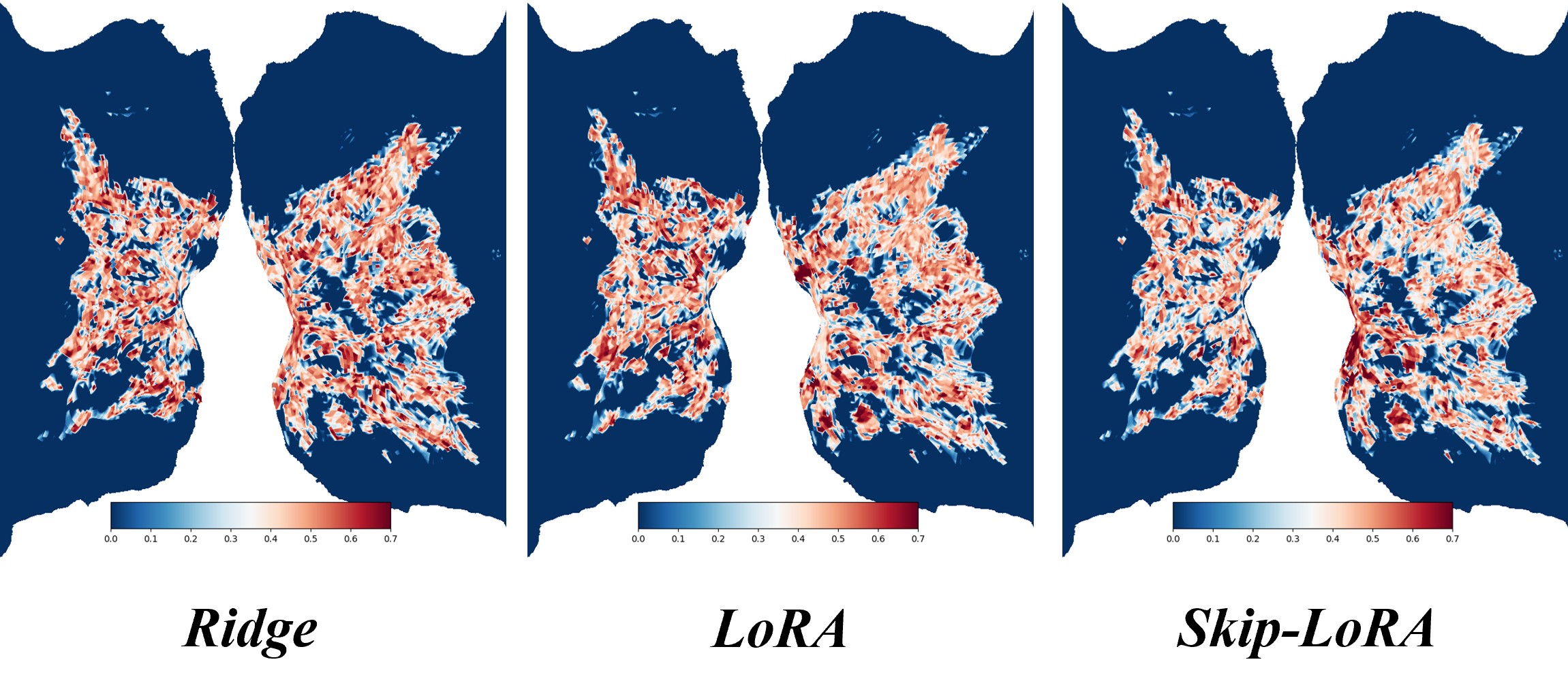}
   \caption{The importance of different ROIs of Subject 7 with
fitted weights of the first layer.}
   \label{subj07}
\end{figure*}

In this section, we visualize the voxel weight maps of the other three subjects in Figure \ref{subj02}, \ref{subj05} and \ref{subj07}, and the templates for the brain region references associated with \textit{nsdgeneral} are shown in Figure \ref{fsaverage}. It can be seen that the content of the visual fingerprint is not quite the same across subjects, but both LoRA and SKip-LoRA have a portion of the primary visual cortex voxels that are significantly more heavily weighted than Ridge. As with Subj01, The visual fingerprint is more heavily weighted on the higher visual cortex for Subj02 and Subj05. However, this phenomenon was not observed for Subj07, which may be the reason why Subj07 performs worse on all the metrics. Further reasons need more explanations from neuroscientific mechanisms.

\end{document}